# SVD Based Image Processing Applications: State of The Art, Contributions and Research Challenges

Rowayda A. Sadek*

Computer Engineering Department, College of Engineering and Technology, Arab Academy for Science
Technology & Maritime Transport (AASTMT), Cairo, Egypt

*Abstract*— Singular Value Decomposition (SVD) has recently emerged as a new paradigm for processing different types of images. SVD is an attractive algebraic transform for image processing applications. The paper proposes an experimental survey for the SVD as an efficient transform in image processing applications. Despite the well-known fact that SVD offers attractive properties in imaging, the exploring of using its properties in various image applications is currently at its infancy. Since the SVD has many attractive properties have not been utilized, this paper contributes in using these generous properties in newly image applications and gives a highly recommendation for more research challenges. In this paper, the SVD properties for images are experimentally presented to be utilized in developing new SVD-based image processing applications. The paper offers survey on the developed SVD based image applications. The paper also proposes some new contributions that were originated from SVD properties analysis in different image processing. The aim of this paper is to provide a better understanding of the SVD in image processing and identify important various applications and open research directions in this increasingly important area; SVD based image processing in the future research.

*Keywords- SVD; Image Processing; Singular Value Decomposition; Perceptual; Forensic.*

I. INTRODUCTION

The SVD is the optimal matrix decomposition in a least square sense that it packs the maximum signal energy into as few coefficients as possible [1,2]. Singular value decomposition (SVD) is a stable and effective method to split the system into a set of linearly independent components, each of them bearing own energy contribution. Singular value decomposition (SVD) is a numerical technique used to diagonalize matrices in numerical analysis [3,4]. SVD is an attractive algebraic transform for image processing, because of its endless advantages, such as maximum energy packing which is usually used in compression [5,6], ability to manipulate the image in base of two distinctive subspaces data and noise subspaces [6,7,8], which is usually uses in noise filtering and also was utilized in watermarking applications [9,6]. Each of these applications exploit key properties of the SVD. Also it is usually used in solving of least squares problem, computing pseudo- inverse of a matrix and multivariate analysis. SVD is robust and reliable orthogonal matrix decomposition methods, which is due to its conceptual and stability reasons becoming more and more popular in signal processing area [3,4]. SVD has the ability to adapt to the variations in local statistics of an image [5]. Many SVD properties are attractive and are still not fully utilized. This paper provides thoroughly experiments for the generous properties of SVD that are not yet totally exploited in digital image processing. The developed SVD based image processing techniques were focused in compression, watermarking and quality measure [3,8,10,11,12]. Experiments in this paper are performed to validate some of will known but unutilized properties of SVD in image processing applications. This paper contributes in utilizing SVD generous properties that are not unexploited in image processing. This paper also introduces new trends and challenges in using SVD in image processing applications. Some of these new trends are well examined experimentally in this paper and validated and others are demonstrated and needs more work to be maturely validated. This paper opens many tracks for future work in using SVD as an imperative tool in signal processing.

Organization of this paper is as follows. Section two introduces the SVD. Section three explores the SVD properties with their examining in image processing. Section four provides the SVD rank approximation and subspaces based image applications. Section five explores SVD singular value based image applications. Section six investigates SVD singular vectors based image applications. Section seven provides SVD based image applications open issues and research trends.

II. SINGULAR VALUE DECOMPOSITION (SVD)

In the linear algebra the SVD is a factorization of a rectangular real or complex matrix analogous to the digonaliztion of symmetric or Hermitian square matrices using a basis of eigenvectors. SVD is a stable and an effective method to split the system into a set of linearly independent components, each of them bearing own energy contribution [1,3]. A digital Image X of size MxN, with M≥N, can be represented by its SVD as follows;

$$[X]_{M}^{N} = {}_{M}[U]^{M} {}_{M}[S]_{M}^{N} {}_{N}[V]^{N^{T}}_{N} \quad (1\text{-}a)$$

$U = [u_1, u_2, \ldots u_m]$,    $V = [v_1, v_2, \ldots v_n]$,

$$S = \begin{bmatrix} \sigma_1 & & & \\ & \sigma_2 & & \\ & & O & \\ & & & \sigma_n \end{bmatrix} \quad (1\text{-}b)$$

Where *U* is an *MxM* orthogonal matrix, *V* is an *NxN* orthogonal matrix, and *S* is an *MxN* matrix with the diagonal elements represents the singular values, $s_i$ of *X*. Using the subscript *T* to denote the transpose of the matrix. The columns of the orthogonal matrix *U* are called the left singular vectors, and the columns of the orthogonal matrix *V* are called the right singular vectors. The left singular vectors (LSCs) of *X* are eigenvectors of $XX^T$ and the right singular vectors (RSCs) of *X* are eigenvectors of $X^TX$. Each singular value (SV) specifies the *luminance* of an image layer while the corresponding pair of singular vectors (SCs) specifies the *geometry* of the image [13]. U and V are unitary orthogonal matrices (the sum of squares of each column is unity and all the columns are uncorrelated) and S is a diagonal matrix (only the leading diagonal has non-zero values) of decreasing singular values. The singular value of each eigenimage is simply its 2-norm. Because SVD maximizes the largest singular values, the first eigenimage is the pattern that accounts for the greatest amount of the variance-covariance structure [3,4].

III. SVD IMAGE PROPERTIES

SVD is robust and reliable orthogonal matrix decomposition method. Due to SVD conceptual and stability reasons, it becomes more and more popular in signal processing area. SVD is an attractive algebraic transform for image processing. SVD has prominent properties in imaging. This section explores the main SVD properties that may be utilized in image processing. Although some SVD properties are fully utilized in image processing, others still needs more investigation and contributed to. Several SVD properties are highly advantageous for images such as; its maximum energy packing, solving of least squares problem, computing pseudo-inverse of a matrix and multivariate analysis [1,2]. A key property of SVD is its relation to the rank of a matrix and its ability to approximate matrices of a given rank. Digital images are often represented by *low ra*nk matrices and, therefore, able to be described by a sum of a relatively small set of eigenimages. This concept rises the manipulating of the signal as two distinct subspaces [3,4]. Some hypotheses will be provided and verified in the following sections. For a complete review, the theoretical SVD related theorems are firstly summarized, and then the practical properties are reviewed associated with some experiments.

- **SVD Subspaces:** SVD is constituted from two orthogonal dominant and subdominant subspaces. This corresponds to partition the M-dimensional vector space into *dominant* and *subdominant subspaces* [1,8]. This attractive property of SVD is utilized in noise filtering and watermarking [7,9].

- **SVD architecture:** For SVD decomposition of an image, singular value (SV) specifies the luminance of an image layer while the corresponding pair singular vectors (SCs) specify the geometry of the image layer. The largest object components in an image found using the SVD generally correspond to eigenimages associated with the largest singular values, while image *noise* corresponds to eigenimages associated with the SVs [3,4]

- **PCA versus SVD:** Principle component analysis (PCA) is also called the Karhunen-Loéve transform (KLT) or the hotelling transform. PCA is used to compute the dominant vectors representing a given data set and provides an optimal basis for minimum mean squared reconstruction of the given data. The computational basis of PCA is the calculation of the SVD of the data matrix, or equivalently the eigenvalues decomposition of the data covariance matrix SVD is closely related to the standard eigenvalues-eigenvector or spectral decomposition of a square matrix X, into VLV', where V is orthogonal, and L are diagonal. In fact U and V of SVD represent the eigenvectors for XX' and X'X respectively. If X is symmetric, the singular values of X are the absolute value of the eigenvalues of X [3,4].

- **SVD Multiresolution:** SVD has the maximum energy packing among the other transforms. In many applications, it is useful to obtain a statistical characterization of an image at several resolutions. SVD decomposes a matrix into orthogonal components with which optimal sub rank approximations may be obtained. With the multiresolution SVD, the following important characteristics of an image may be measured, at each of the several level of resolution: isotropy, spercity of principal components, self-similarity under scaling, and resolution of the mean squared error into meaningful components. [5,14].

- **SVD Oriented Energy:** In SVD analysis of oriented energy both rank of the problem and signal space orientation can be determined. SVD is a stable and effective method to split the system into a set of linearly independent components, each of them bearing its own energy contribution. SVD is represented as a linear combination of its principle components, a few dominate components are bearing the rank of the observed system and can be severely reduced. The oriented energy concept is an effective tool to separate signals from different sources, or to select signal subspaces of maximal signal activity and integrity [1, 15]. Recall that the singular values represent the square root of the energy in corresponding principal direction. The dominant direction could equal to the first singular vector $V_1$ from the SVD decomposition. Accuracy of dominance of the estimate could be measured by obtaining the difference or normalized difference between the first two SVs [16].

Some of the SVD properties are not fully utilized in image processing applications. These unused properties will be experimentally conducted in the following sections for more convenient utilization of these properties in various images processing application. Much research work needs to be done in utilizing this generous transform.

IV. SVD-BASED ORTHOGONAL SUBSPACES AND RANK APPROXIMATION

SVD decomposes a matrix into orthogonal components with which optimal sub rank approximations may be obtained.

The largest object components in an image found using the SVD generally correspond to eigenimages associated with the largest singular values, while image *noise* corresponds to eigenimages associated with the smallest singular values. The SVD is used to approximate the matrix decomposing the data into an optimal estimate of the signal and the noise components. This property is one of the most important properties of the SVD decomposition in noise filtering, compression and forensic which could also treated as adding noise in a proper detectable way.

### A. Rank Approximation

SVD can offer low rank approximation which could be optimal sub rank approximations by considering the largest singular value that pack most of the energy contained in the image [5,14]. SVD shows how a matrix may be represented by a sum of rank-one matrices. The approximation a matrix X can be represented as truncated matrix $X_k$ which has a specific rank k. The usage of SVD for matrix *approximation* has a number of practical advantages, such as storing the approximation $X_k$ of a matrix instead of the whole matrix X as the case in image compression and recently watermarking applications. Assume $X \in R^{m \times n}$. Let p =min(m,n), k≤p be the number of nonzero singular values of X. X matrix can be expressed as

$$X = \sum_{i=1}^{k} s_i u_i v_i^T \approx s_1 u_1 v_1^T + s_2 u_2 v_2^T + \ldots + s_k u_k v_k^T \quad (2)$$

i.e., X is the sum of k rank-one matrices. The partial sum captures as much of the "energy" of X as possible by a matrix of at most rank r. In this case, "energy" is defined by the 2-norm or the Frobenius norm. Each outer product $(u_i v_i^T)$ is a simple matrix of rank "1"and can be stored in M+N numbers, versus M*N of the original matrix. For truncated SVD transformation with rank k, storage has **(m+n+1)*k**. Figure (1) shows an example for the SVD truncation for rank k =20.

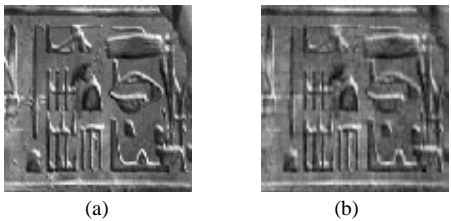

(a)             (b)

Figure 1. Truncated SVD (a) Original (b) Truncated SVD

### B. Orthogonal Subspaces

The original data matrix X is decomposed into the orthogonal dominant components $US_k V^T$, which is the rank k subspace corresponding to the signal subspace and $US_{n-k} V^T$, which corresponds to the orthogonal subdominant subspace that defines the noise components. In other words, SVD has orthogonal Subspaces; dominant and subdominant subspaces. SVD provides an explicit representation of the range and null space of a matrix X. The right singular vectors corresponding to vanishing singular values of X span the null space of X. The left singular vectors corresponding to the non-zero singular values of X span the range of X. As a consequence, the rank of X equals the number of non-zero singular values which is the same as the number of non-zero diagonal elements in S. This is corresponding to partition the M-dimensional vector space (of the mapping defined by X) into *dominant* and *subdominant subspaces* [8]. Figure (2) shows image data dominant subspace with the image truncated to k=30 SVD components, and its subdominant; noise subspace. The SVD offers a good and efficient way to determine the rank(X), orthonormal basis for range(X), null(X), $\|X\|_2$, $\|X\|_{Fro}$ and optimal low-rank approximations to X in $\|\cdot\|_2$ or $\|\cdot\|_F$, etc.

rank(X) = r = the number of nonzero singular values.

range(X) = span($u_1, u_2, \ldots, u_r$)

null(X) = span($v_{r+1}, v_{r+2}, \ldots, v_n$)

This subspaces SVD property that offers splitting the image space into two distinct subspaces, the signal and the noise, triggers proposing a contribution in watermarking application in this paper. The contribution utilizes the resemblance between the SVD domain with any noisy image (signal subspace + noise subspace), or the watermarked image form (image signal+watermark signal).

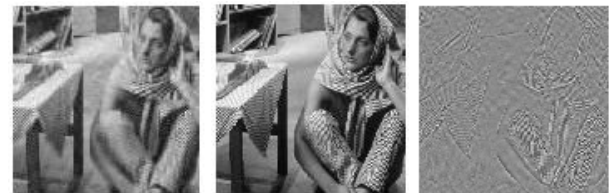

(a)        (b)        (c)

Figure 2. SVD subspaces (a) Original Image (b) Image Data subspace (c) Noise subspace

### C. Image Denoising

SVD has the ability to manipulate the image in the base of two distinctive data and noise subspaces which is usually used in noise filtering and also could be utilized in watermarking [7,9]. Since the generic *noise signal filtering* model assumes the noise can be separated from the data, SVD locates the noise component in a subspace orthogonal to the data signal subspace. Therefore, SVD is used to approximate the matrix decomposing the data into an optimal estimate of the signal and the noise components. Image noise manifests itself as an increased "spatial activity" in spatial domain that guides to increasing the smaller singular values in SVD domain. As there is an added noise, s*ingular values* are non-uniformly increased (although some may decrease) by amount that depends on the image and the noise statistics, the medium values are increased by largest amount, although smallest singular values have the largest relative change. This depicted function will be more or less skewed for different images and noise types. For *Singular vectors* which are noised, it is hard, if not impossible to analytically describe influence of noise on noised singular vectors. Singular vectors that correspond to smaller singular values are much more perturbed. Degradation from noising of singular vectors is much bigger than that caused by increased singular values. Incautious changes in singular vectors can produce catastrophic changes in images. This is the reason why the filtering operations are limited to slight filtering of noise in singular vectors [7]. Based on the fact of non-uniformly affecting the SVs and SCs by noise based on its statistics, smallest SVs and faster changing singular vectors which correspond to higher r values are highly affected with the noise compared to the larger SVs and their corresponding SCs. [7,8]. This intuitive explanation is validated experimentally as shown in figure (3).

Figure (3) shows the 2-dimensional representation of the left and right SCs.

The slower changing waveform of the former SCs is versus the faster changing of latter SCs. Figure (4) shows the orthogonality of the different subspaces by carrying out correlation between different slices. Figure (5) shows the SVD based denoising process by considering the first 30 eigenimages as image data subspace and the reminder as the noise subspace. By removing the noise subspace, image displayed in figure(5b) represents the image after noise removal.

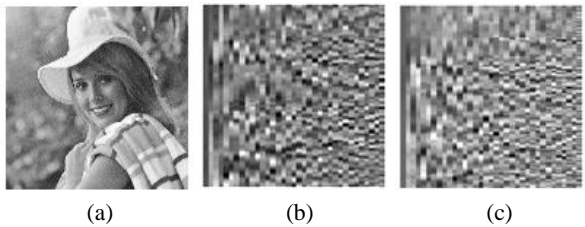

(a)    (b)    (c)

Figure 3.  2D Representation of SCs: (a) Original Image (b) Left SCs; U  (c) Right SCs; V

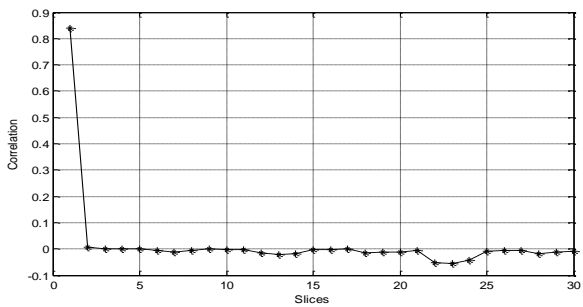

Figure 4.  Correlation is carried out between different subspaces (slices)

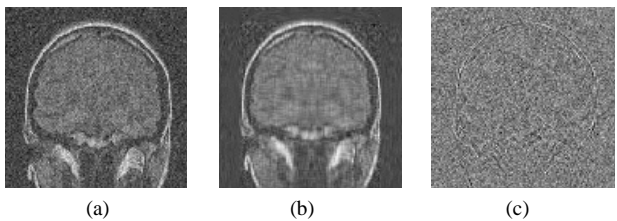

(a)    (b)    (c)

Figure 5.  SVD Denoising (a) Original Noisy MRI Image (b) Image Data subspace (c) Noise subspace

### D. Image Compression

SVD with the maximum energy packing property is usually used in compression. As mentioned above, SVD decomposes a matrix into orthogonal components with which optimal sub rank approximations may be obtained [5, 14].

As illustrated in equation 2, truncated SVD transformation with rank r may offer significant savings in storage over storing the whole matrix with accepted quality. Figure (6) shows the bock diagram of the SVD based compression.

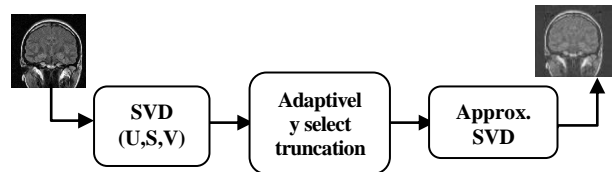

Figure 6.  SVD based Compression

Compression ratio can be calculated as follows;

$$R = \frac{nk + k + mk}{nm} * 100 \qquad (3)$$

Where R is the compression percentage, k is the chosen rank for truncation; m and n are the number of rows and columns in the image respectively. R for the truncated image shown in figure (1) is 15.65 and for the one located in figure (2) are 23.48. Figure (7) shows compressed images with different chosen ranks for truncation that result in different compression ratios. Table 1 illustrates the different truncation levels k used for compressing image shown in figure (7) and the resultant compression ratio for each truncation level. Peak Signal to Noise Ratio (PSNR) is also illustrated in the table 1 corresponding to the different compression ratios to offer objective quality measure

TABLE 1: COMPRESSION VS. PSNR

| Number of truncated levels "k" | Compression "R" | PSNR |
|---|---|---|
| 90 | 70.4498 | 37.7018 |
| 80 | 62.6221 | 36.0502 |
| 60 | 46.9666 | 32.7251 |
| 40 | 31.311 | 32.7251 |
| 20 | 15.6555 | 24.2296 |
| 10 | 7.8278 | 21.3255 |

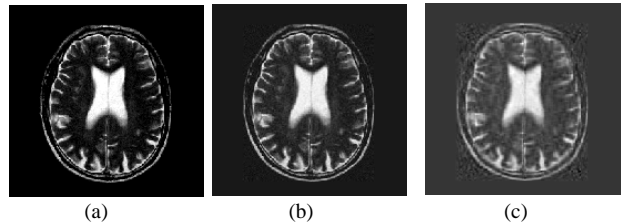

(a)    (b)    (c)

Figure 7.  SVD Based Compression (a) Original (b) Compression 47% (truncation to k=60) (c) Compression 16%  (truncation to k=20)

### E. Image Forensic

For the current digital age, digital forensic research becomes imperative. Counterfeiting and falsifying digital data or digital evidence with the goal of making illegal profits or bypassing laws is main objective for the attackers [15]. The forensic research focuses in many tracks; steganography, watermarking, authentication, labeling, captioning, etc. Many applications were developed to satisfy consumer requirements such as labeling, fingerprinting, authentication, copy control for DVD, hardware/ software watermarking, executables watermarks, signaling (signal information for automatic counting) for propose of broadcast monitoring count [15].

The SVD packs the maximum signal energy into as few coefficients. It has the ability to adapt to the variations in local statistics of an image. However, SVD is an image adaptive transform; the transform itself needs to be represented in order to recover the data. Most of the developed SVD based watermarking techniques utilizes the stability of singular values (SVs) that specifies the luminance (energy) of the image layer [13,18]. That is why slight variations of singular values could not influence remarkably on the cover image quality. Developed SVD based techniques either used the largest SVs [13,19] or the lowest SVs to embed the watermark components either additively [18] or by using quantization [20]. D. Chandra [18] additively embedded the scaled singular values of watermark into the *singular values* of the host image *X* as described above.

### *The Proposed Perceptual Forensic Technique*

A new perceptual forensic SVD based approach which is based on global SVD (GSVD) is proposed in this paper,. This technique is developed to be private (non-blind) forensic tool. The proposed forensic tool is based on efficient additively embedding the optimal watermark data subspace into the host less significant subspace (noise subspace). This forensic tool can be utilized in all the forensic applications with some kind of adaptation in the embedding region based on the required robustness. Although many SVD based embedding techniques for many forensic purposes are carried out additively in singular values, they considered scaled addition without considering the wide range of singular values. The proposed scaled addition process that carried out for the SVs is treated differently because of the wide range of the SVs sequence which required to be flatted for convenient addition. Embedding is carried out by getting the SVD for image X and watermark W as follows in Eq. (4-a) and Eq. (4-b). The scaled addition is as in Eq. (4-c). Finally watermarked image "*Y*" is reconstructed from the modified singular values $S_m$ of the host image as in Eq.(4-d).

$$X = U_h S_h V_h^T \qquad (4\text{-a})$$

$$W = U_w S_w V_w^T \qquad (4\text{-b})$$

$$S_m(i) = \begin{cases} S_h(i) + \alpha * \log(S_w(q)) & \text{if } M-k < i < M, \quad 1 \le q \le k \\ S_h(i) & \text{Otherwise} \end{cases} \qquad (4\text{-c})$$

$$Y = U_m S_m V_m^T \qquad (4\text{-d})$$

Where $S_m$ $S_h$ and $S_w$ are the singular values for the modified media, host media and embedded data respectively. $\alpha$ is a scaling factor which is adjustable by user to increase (or decrease) the protected image fidelity and decrease (or increase) the security of watermark protection, and robustness at the same time. "k" is user defined, and could be chosen adaptively based on the energy distribution in both of the host and the embedded data (watermark). k represents the size of *range*(embedded data) and *null*(host data). Since the SVs has wide range, they should be treated differently to avoid the abrupt change in the SVs sequence of the resultant watermarked media which sure will give sever degradation. Therefore, log transformation is proposed to solve this problem by flatten the range of watermark SVs in order to be imperceptibly embedding.

The detection is non-blind. The detection process is performed as the embedding but in the reverse ordering. Obtain SVD of the watermarked image "Y" and "X" as in Eq. (5-a,b). Then obtain the extracted watermark components $S'_w$ as shown in Eq.(5-c). Finally, construct the reconstructed watermark W' by obtaining the inverse of the SVD.

$$Y = U_m S_m V_m^T \qquad (5\text{-a})$$

$$X = U_h S_h V_h^T \qquad (5\text{-b})$$

$$S'_w(i) = Exp((S(i)-S_h(i))/\alpha) \quad for\ M\text{-}k<i<M \qquad (5\text{-c})$$

$$W' = U_w S'_w V_w^T \qquad (5\text{-d})$$

Fidelity measure by using Normalized Mean Square Error (NMSE) or Peak Signal to Noise Ratio (PSNR) is used to examine perceptually of the watermarked image. The security of the embedding process lays on many parameters, number of selected layers used in truncation as Truncated SVD (TSVD) for the watermark efficient compression, the starting column in host SVD components that were used for embedding. Experimentally examining this proposed forensic technique is carried out as compared to commonly used developed technique [19].

Figure (8) shows the watermarked image by using proposed forensic technique (as in Eq.4) compared to the already developed Chandra's scheme [19]. Figure (8a) shows the effect of logarithmic transformation in the SVs sequence range. Chandra's scheme that used constant scaling factor $\alpha$ to scale the wide range of SVs produced anomalous change (zoomed part) in the produced watermarked SVs sequence compared to the original SVs sequence while the proposed technique produces SVs smoothed sequence much similar to the original SVs sequence.

Figure (8c, d) show the watermarked images by using scaled addition of the SVs [sv01] and by using the proposed logarithmic scaled of SVs addition with using the same scaling factor ($\alpha$=0.2). Objective quality measure by using NMSE values for developed and proposed techniques are 0.0223 and 8.8058e-009 respectively. Subjective quality measure shows the high quality of the resultant image from the proposed technique compared to the developed one. Figure (9) also examines the transparency by using a kind of high quality images; medical images.

Both objectively and subjectively measures proves the superiority of the proposed technique in transparency. NMSE values for developed and proposed techniques are 0.0304 and 8.3666e-008 respectively.

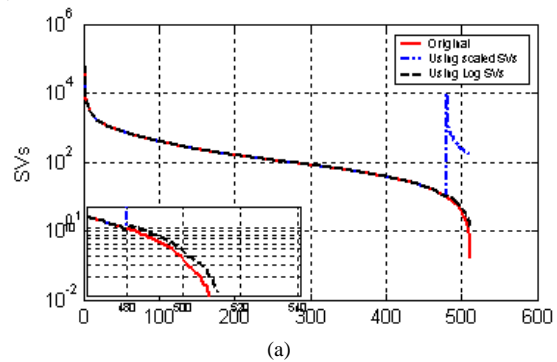

(a)

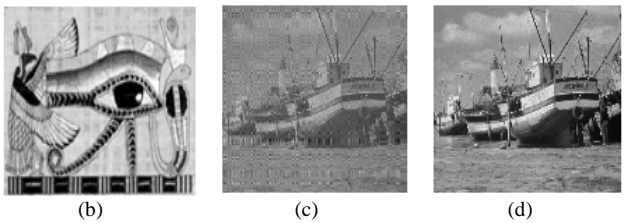

(b)     (c)     (d)

Figure 8. Effect of logarithmic transformation on SVs range (a) SVs sequences of original, scaled and logged one. (b)Watermark (c)Watermarked image using scaled addition of watermark SVs (d) Watermarked image using scaled addition of log of watermark SVs

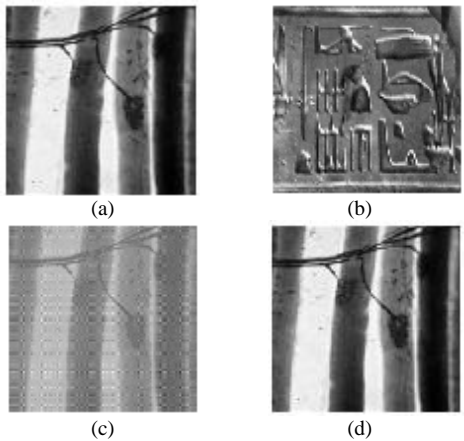

(a)     (b)
(c)     (d)

Figure 9. Perceptual SVD forensic: (a) Original (b) Watermark (c)Watermarked image using scaled addition of watermark SVs (d) Watermarked image using scaled addition of log of watermark SVs

## V. SVD SINGULAR VALUES CHARACTERISTICS

Since each singular value of image SVD specifies the luminance (energy) of the image layer and respective pair of singular vectors specifies image topology (geometry). That is why slight variations of singular values could not influence remarkably on the cover image quality [13]. Singular values distribution and their decaying rate are valuable characteristics.

### A. Singular Values Distribution

Since SVs represent the luminance, SVs of two visual distinct images may be almost similar and the corresponding U and V of their SVD are different because they are representing image structure. This fact was examined and proved [15]. Figure(10) shows the closeness among the SVs of two different images. Figure (11) demonstrates the reconstruction of the image from its truncated 30 singular vectors and singular values of the image itself and the two different images used in the previous figure with NMSE; 0.0046, 0.0086 and 0.0292 respectively. This makes hiding any data in SVs values is vulnerable to illumination attack and fragile to any illumination processing [15]. This valuable feature could be utilized with more research in the application such as; Stegano-analysis for SVD based steganography and forensic techniques, Illumination attacking for SVD based forensic techniques and image enhancement by using selected SVs of a light image analogy with the histogram matching [15].

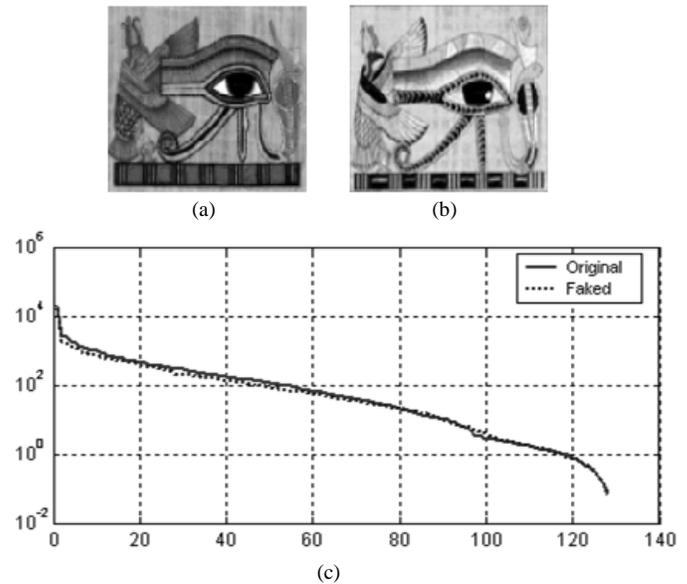

(a)     (b)

(c)

Figure 10. Similarity of SVs (a) Original Image (b) Faked image (c) SVs of both of (a) and (b).

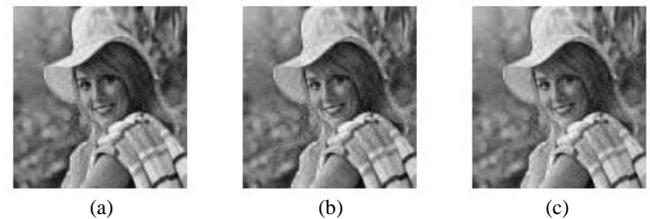

(a)     (b)     (c)

Figure 11. Reconstructed Image From 30 SVD truncated components (a) its SVs (b) SVs of figure(10a) (c) SVs of figure(10b)

### B. Singular Values Decaying

s*ingular values* are non-uniformly increased (although some may decrease) by amount that depends on the image and the noise statistics, the medium values are increased by largest amount, although smallest singular values have the largest relative change. This depicted function will be more or less skewed for different images and noise types [7, 9].

Relying on the fact that says "Smooth images have SVs with rapid decaying versus slow decaying of SVs of randomly images", slope of SVs could be used as a roughness measure. Figure (12) shows the rapid decaying of singular values of smooth image versus those of the noisy image.

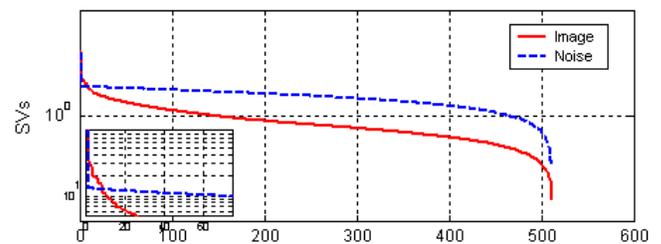

Figure 12. Rate of SVs decaying

## C. Image Roughness Measure

Roughness measure is inversely proportional with the decaying rate. Roughness measure could be used in the application of perceptual based nature that considers the human visual system (HVS) such as perceptual coding and perceptual data embedding. Figure (13) shows the rapid decaying of singular values of smooth low pass filtered image versus those of the original image without filtering. The payload capacity of any host image to hide data could also be measured also based on roughness measure. Since the condition number (CN) is the measure of linear independence between the column vectors of the matrix X. The CN is the ratio between largest and smallest SVs. The CN value could be used for the block based processing by finding the CN for each block as follows;

$$CN_B = \frac{S_{B\max}}{S_{B\min}} \quad (6)$$

Sensitivity to noise increases with the increasing of the condition number. Lower CN values correspond to random images which usually bear more imperceptible data embedding. Conversely, the higher CN correspond to smooth images which don't bear embedding data, Smooth blocks ( high CN till ∞) and rough detailed blocks (with low CN till one for random block). $Rf_B$ is the roughness measure in a block B.

$$Rf_B = d.\frac{1}{CN_B} \quad (7)$$

d is a constant. $Rf_B$ ranges from d for highly roughness block to 0 for the completely homogenous smoothly block. This valuable feature could be utilized with more research in the adaptive block based compression.

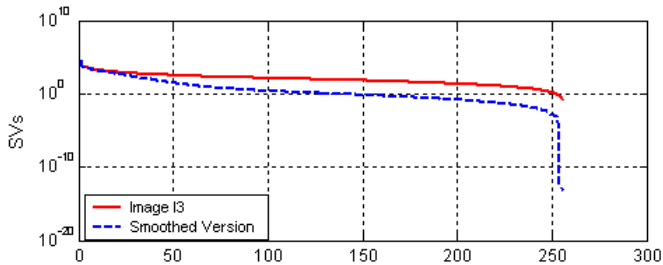

Figure 13. LPF Effect on SVs of an image and its smoothed version

## VI. SVD SINGULAR VECTORS CHARACTERISTICS

Since singular vectors specifiy image geometry, two visual distinct images may have singular values but the U and V of the SVD are different [15]. First singular vectors are slower changing waveforms, they describe global shape of the scene in the vertical and horizontal direction. This was experimentally examined in figure (3). One of the important applications of SVD is the analysis of oriented energy. SVD is a stable and effective method to split the system into a set of linearly independent components, each of them bearing own energy contribution. Thus signal space orientation can be determined. The oriented energy concept is an effective tool to separate signals from different sources, to separate fill noisy signals or to select signal subspaces of maximal signal activity [1,2]. The *norm* of a matrix is a scalar that gives some measure of the magnitude of the elements of the matrix [18,20].

### A. Main dominant directions in the image structure.

For SVD, each direction of the critical oriented energy is generated by right singular vector "V" with the critical energy equal to the corresponding singular value squared. The left singular vectors "U" represent each sample's contribution to the corresponding principle direction. It is well known in the earlier works that the singular values can be seen as the measure of *concentration* around the principal axes. The image orientation can be obtained from the first singular vector (note that the gradient vector are orthogonal to the image orientation we seek, so after obtaining the principal direction of the gradient vectors, we need to rotate by $\pi/2$ to get the orientation we want) [16]. Singular values represent the square root of the energy in corresponding principal direction. The dominant direction could equal to the first singular vector (SC); $V_1$ from the SVD decomposition. Accuracy of dominance of the estimate could be measured by obtaining the difference or normalized difference between the first two SVs [16]. Figure (14) shows three different images; brain, horizontal rays and vertical rays. Figure (14) also displays the SCs; U and V for all the images as well as their SVs in a graph. Graph of the SVs shows the difference in convergence to a rank.

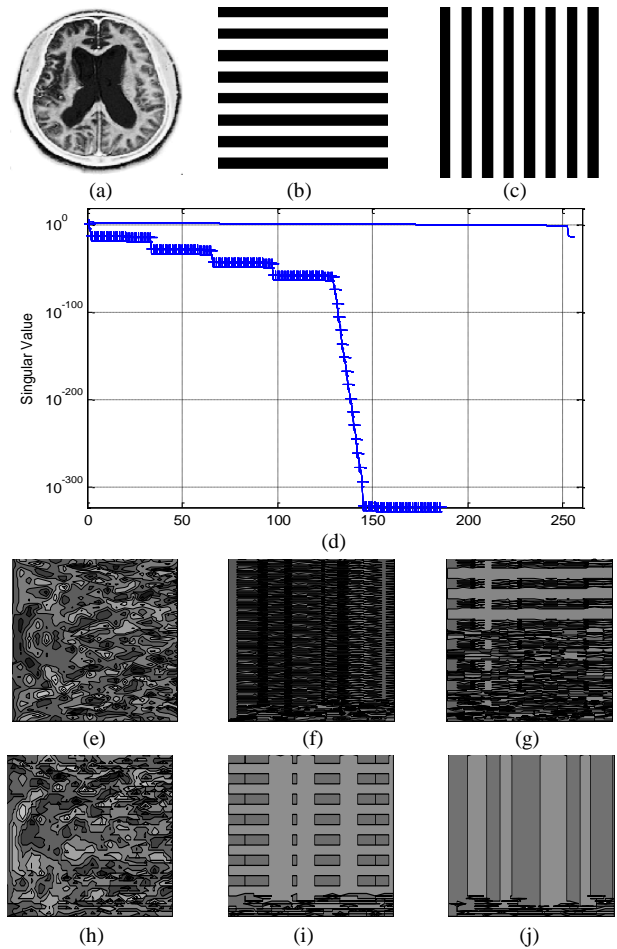

Figure 14. Figure 14 SVD orientation (a-c) brain, horizontal and vertical images respectively (d) Their SVs. (e-g) V for each image respectively (h-j) U for each image respectively.

*B. Frobenius based Energy Truncation*

The *norm* of a matrix is a scalar that gives some measure of the magnitude of the elements of the matrix [18,20]. For n-element vector A, Norm is equivalent to Euclidean length therefore Norm-2 sometimes called Euclidean Norm. Euclidean length is the square root of the summation of the squared vector elements

$$\text{Norm}(A) = \sqrt{\sum_{i=1}^{n}(A_i)^2} \quad (8)$$

where $A_i$, is n-element vector $i = 1,..,n$ are the components of vector $V^\lambda$. This is also called Euclidean norm or Norm-2 which is equivalent the largest singular value that results from the SVD of the vector A.

$$\|A\|_2 = \sigma_1 \quad (9)$$

The Frobenius-norm of the mxn matrix A which is equivalent to

$$\text{Norm}_F(A) = \sqrt{\sum_{i=1}^{m}\sum_{j=1}^{n} \text{diag}(A'*A)} = \sqrt{\sum_{i=1}^{m}\sum_{j=1}^{n}|A|^2} \quad (10)$$

This Frobenius norm can be calculated directly as follows;

$$\|A\|_F = \sqrt{\sigma_1^2 + \sigma_2^2 + \Lambda + \sigma_n^2} = \sqrt{\sum_{i=1}^{n}\sigma_i^2} \quad (11)$$

A pre-selected number "k" of SVD components (k image layers) is to be truncated for efficient truncation in different applications. This number of image layers could be selected based on the energy content instead of using hard-threshold value. Frobenius norm could be used as the base of content energy measurement, the required specified contained energy; $E_k$ could be represented as follows;

$$E_k = \frac{\|A_k\|_F}{\|A\|_F} \quad (12)$$

where A is the image and the $A_k$ is the approximated or truncated image at rank "k". Truncated layers could be specified by specifying the number of layers "k" required to have 90% of the host ($E_k \geq 0.9$)

*C. Frobenius based Error Truncation*

Frobenius error agrees with the error based on visual perception, thus a threshold to preserve the required quality can be controlled by using Frobenius norm; by specifying an error threshold to avoid exceed it

$$\varepsilon_F = \frac{\|A - A_k\|_F}{\|A\|_F} \quad (13)$$

$\varepsilon_F$ is the Frobenius error that is calculated from the Frobenius norm of the difference between the image A and its truncated version $A_k$ and normalized with the Frobenius norm of the image A. Frobenius norm can be used to check the error threshold. Find the needed rank that bound the relative error by controlling the Frobenius norm to be less than a predefined bounded threshold. Simply proper "k" number could be selected to satisfy a predefined constraint either on Frobenius energy or Frobenius norms error.

*D. SVD-based Payload Capacity Measure*

SVD transformation of an image may be utilized to give a clue about the image nature, or may be used in a HVS based classification process for image blocks roughness. Image payload capacity and permissible perceptual compression could be achieved by many SVD based methods; as Frobenius energy or Frobenius norms error. The Frobenius based energy of first k values is high means that the image is more smooth and hence it has low capacity for data embedding and this can be proved by considering Eq.(10). Conversely, the detailed image will have less Frobenius energy for the same number layers "k", compared to the higher Frobenius energy of smoothed image. On the other hand, the sensitivity to noise is increased (low capacity) for smooth image and decreased (high capacity) for rough images. Therefore, the capacity of the image to bear hidden information or to bear more compression without perceptually noticeable effects is increasing for rough or high detailed image and vice versa. Therefore, the chosen suitable number of "k" layers can lead to a certain predefined quality PSNR. Figure(15) shows the capacity of the image in carrying hidden data or has compression in adaptively block based manner. The block based capacity calculation uses 16x16 block size.

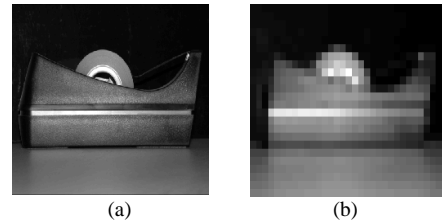

Figure 15. Block based Capacity calculation (a) Original (b) Capacity

VII. CONCLUSION AND OPEN ISSUES AND RESEARCH TRENDS

Despite the attention it has received in the last years, SVD in image processing is still in its infancy. Many SVD characteristics are still unutilized in image processing. This paper proposed a through practical survey for SVD characteristics in various developed image processing approaches. The paper also proposed contribution in using unused SVD characteristics in novel approaches such as adaptive block based compression, perceptual multiple watermarking, image capacity for hiding information, roughness measure, etc, All these contributions were experimentally examined and gave promising results compared to developed ones. The main contributions in this paper are a novel perceptual image forensic technique, a new prospective vision in utilizing the SVD Properties, reviewing and experimental valuation of the developed SVD based application such as denoising, compression, a new block based roughness measure for application such as perceptual progressive compression as well as perceptual progressive data hiding. Image denoising and compression were thoroughly examined and provided good results although they are image dependent. Perceptual fragile forensic tool gives highly promising results compared to the commonly used SVD based tool. Energy based truncation and error based truncation as well as the roughness measures are promising in many application.

The paper also suggests some open research issues which require more research and development such as calculating the block based dominant orientation, adaptively image fusion, block based robust forensic, etc. On the other hand, more utilization for proposed valuable feature of closeness between the SVs of different images used in applications such as; Stegano-analysis for SVD based steganography and forensic techniques, Illumination attacking for SVD based forensic techniques, image enhancement by using SVs matching in analogy with the histogram matching, etc. The proposed SVD based roughness measure could be also utilized in the application such as; adaptive block based compression, payload capacity measure for images in forensic tool, etc.

AUTHOR PROFILE

**Rowayda A. Sadek** received B.Sc., MSc and PhD degrees from Alexandria University. She is currently an Assistant Professor in Computer Engineering Department, Faculty of Engineering, Arab Academy on Technology and Marine. She is on temporary leave from Helwan University. Her research interests are in Multimedia Processing, Networking and Security. Dr. Rowayda is a member of IEEE.